\documentclass[10pt,twocolumn,letterpaper]{article}

\usepackage{cvpr}              % To produce the CAMERA-READY version
\usepackage{booktabs}
\usepackage{amssymb}
\definecolor{cvprblue}{rgb}{0.21,0.49,0.74}
\usepackage[pagebackref,breaklinks,colorlinks,allcolors=cvprblue]{hyperref}

\def\confName{CVPR}
\def\confYear{2027}

\title{EG360: Efficient and Generalizable $360^\circ$ Depth Estimation via Adaptive Panorama-Perspective Distillation and Spherical Cross-Level Feature Fusion}

\author{
Laiyan Ding\\
CUHKSZ\\
\and
Chongxi Meng\\
Tongji University\\
\and
Rui Huang\\
CUHKSZ\\
}

\begin{document}
\maketitle
\begin{abstract}
Generalized panorama depth estimation has witnessed great advances in recent years. These methods typically achieve robust zero-shot generalization performance using large pretrained models, but their high computational cost makes real-time inference on embedded platforms challenging. In contrast, previous lightweight panorama models have mostly been developed for single-domain evaluations and cannot generalize well due to domain shifts. To reduce the gap between these two types of models, we introduce EG360, an efficient and generalizable panorama depth estimation model. Specifically, building on a lightweight depth model, we design an Adaptive Panorama-Perspective Distillation (APPD) strategy guided by scene geometry, along with a Spherical Cross-Level Feature Fusion (SCFF) module for global semantic perception. As a result, with only 6.593M parameters after reparameterization, EG360 achieves strong zero-shot performance on real-world benchmarks and runs at 35 FPS on a Jetson Orin NX 16GB in MAXN mode, whereas a previous ViT-small model runs at less than 2 FPS. Our code will also be made public for future research.

\end{abstract}    
\section{Introduction}
\label{sec:intro}

In recent years, there has been a wide range of work that addresses challenges in processing panoramic images for various computer vision tasks, including depth estimation \cite{lin2026depth,jiang2021unifuse,bozic2026unified}, segmentation \cite{zhang2024goodsam,zhang2022bending,zhong2025omnisam}, and tracking \cite{he2021know,luo2025omnidirectional,xu2025360vots}. Among these, monocular panorama depth estimation is an important task that infers the 3D geometry of the corresponding scene from a 360 degree image, which can be useful for other applications such as virtual reality \cite{asija20243d, wang2024perf} and autonomous driving \cite{teng2026hallucinating, payen2018eliminating}.

\begin{figure}[t]
    \centering
    \includegraphics[width=\linewidth]{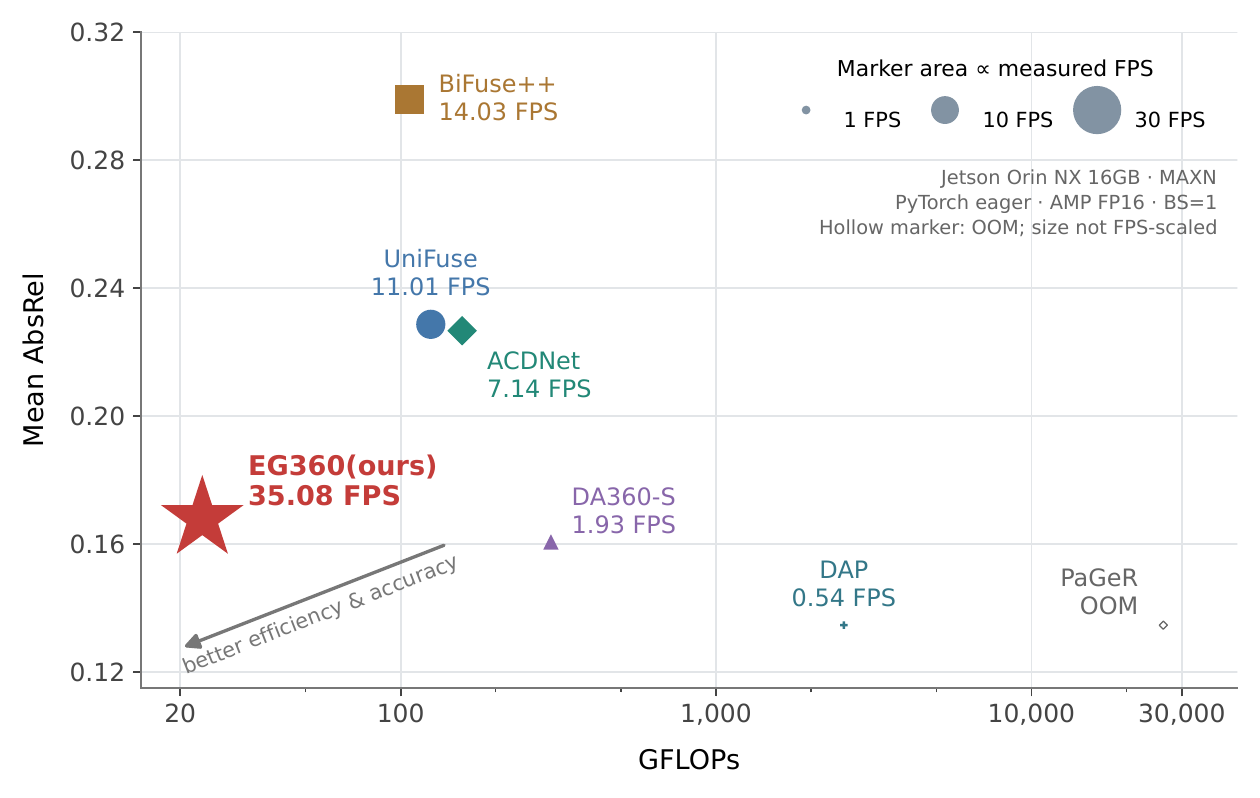}
    \caption{Accuracy–computation trade-off for panoramic depth estimation. Mean AbsRel is averaged equally over Matterport3D, Stanford2D3D, and Metropolis. Filled marker area is positively proportional to measured inference FPS on a Jetson Orin NX 16GB (MAXN mode, PyTorch eager, AMP FP16, batch size 1). EG360 uses reparameterization for inference. PaGeR ran into out-of-memory (OOM) and is represented by a fixed-size hollow diamond.}
    \label{fig:compare_methods}
\end{figure}

In the domain of monocular panorama depth estimation, some generalizable methods \cite{wang2026depthmaster, lin2026depth} that can perform robust zero-shot predictions have emerged. Though they can achieve good performance, they tend to be based on large foundation models \cite{oquab2023dinov2,dosovitskiy2020image}, which are hard to meet real-time requirements on mobile platforms, e.g., Jetson Orin NX, an embedded AI computing platform for robotics. As shown in Fig. \ref{fig:compare_methods}, the DAP \cite{lin2026depth} model, which is built on top of a ViT-Large model \cite{simeoni2025dinov3,dosovitskiy2020image}, obtains an average AbsRel of 0.135 on Matterport3D \cite{chang2017matterport3d}, Stanford2D3D \cite{armeni2017joint} and Metropolis \cite{mapillary2024metropolis}. However, it can only run at 0.54 FPS on a Jetson Orin NX 16GB in MAXN mode. In contrast, previous lightweight models \cite{wang2020bifuse, jiang2021unifuse, zhuang2022acdnet} focus on training and testing within the same domain and show limited cross-domain generalization performance. As shown in Fig. \ref{fig:compare_methods}, UniFuse \cite{jiang2021unifuse} achieves only an average AbsRel of 0.229 and runs at 11 FPS. Thus, current panorama depth estimation models fail to deliver depth estimation results with a plausible trade-off between quality and efficiency.

We argue that to achieve both efficiency and generalization, exposing the model to more real-world data via distillation and allowing it to capture semantic information efficiently are necessary. Concretely, we propose an Adaptive Panorama-Perspective Distillation (APPD) strategy to exploit panorama and perspective teachers jointly. Whereas Depth Anywhere \cite{wang2024depth} and PanDA \cite{cao2025panda} rely on perspective and panoramic teachers, respectively, we combine their complementary supervision. Table \ref{tab:teacher_comparison} shows that DAP \cite{lin2026depth} performs better on Stanford2D3D, which is indoor, while MoGe2 \cite{wang2026moge} outperforms on Metropolis, which is outdoor. Their relative usefulness may also vary across views: a perspective crop dominated by a wall may provide limited cues for estimating absolute distance, whereas a panorama teacher can use the surrounding contexts. Motivated by this, APPD uses the proposed normalized plane-fitting residuals to dynamically adaptively weigh the two teacher losses, assigning more weight to the panorama teacher for more planar views. Additionally, to aggregate global contexts while accounting for spherical geometry, we introduce Spherical Cross-Level Feature Fusion (SCFF). SCFF uses cross-level attention \cite{vaswani2017attention} between the two coarsest encoder feature maps, with the finer map querying the coarser one. Attention combines feature similarity with a spherical direction bias that reflects the spatial proximity on the sphere.

Combining these two proposed techniques, we present EG360, a lightweight monocular panorama depth estimation model that can perform robust zero-shot predictions. Training on abundant labeled data \cite{zheng2020structured3d, wang2020tartanair, songpvdepth} together with unlabeled data \cite{cruz2021zillow, zhou2020holicity}, EG360, comprising 6.593M parameters and costing 23.492 GFLOPs after structural reparameterization \cite{ding2021repvgg}, can achieve an average AbsRel of 0.168  and run on Orin NX at 35 FPS. It not only runs faster than previous lightweight models but also narrows the gap between lightweight and foundation models.

In summary, our contributions are as follows:
\begin{itemize}
    \item We propose Adaptive Panorama--Perspective Distillation (APPD), which uses normalized plane-fitting residuals to weigh panorama and perspective teacher supervision for unlabeled data adaptively.

    \item We introduce Spherical Cross-Level Feature Fusion (SCFF), which combines cross-level attention with a spherical direction bias to aggregate global context between the two coarsest feature maps.

    \item We develop EG360, a lightweight model with 6.593M parameters and 23.492 GFLOPs, and demonstrate its zero-shot performance on real-world benchmarks.
\end{itemize}

\section{Related Works}
\label{sec:related_works}

\subsection{In-Domain Panorama Depth Estimation} Depth estimation for panorama images presents unique challenges due to the equirectangular projection (ERP) and resulting distortions. Some works directly work on the ERP image and propose different techniques to tackle the distortion. OmniDepth \cite{zioulis2018omnidepth} uses both rectangular and square CNN filters. ACDNet \cite{zhuang2022acdnet} combines convolution kernels with different dilations with adaptive channel-wise fusions. PanoFormer \cite{shen2022panoformer} and EGFormer \cite{yun2023egformer} propose panorama-specific transformer blocks to adapt widely used transformer \cite{vaswani2017attention} blocks to the panorama domain. Some other works use additional distortion-free formats for 360 images to mitigate the negative effects of distortions. BiFuse \cite{wang2020bifuse}, BiFuse++ \cite{wang2022bifuse++}, and UniFuse \cite{jiang2021unifuse} use cubemap (CP) images and fuse the features from CP and ERP images adaptively for better performance than using ERP or CP alone. HRDFuse \cite{ai2023hrdfuse} enhances depth prediction by collaboratively learning from ERP and tangent projections. In addition, $S^2$Net \cite{li2023mathcal} moves to the spherical surface generated from the ERP image with the HEALPix sampling \cite{gorski2005healpix}. Nevertheless, these methods are typically trained and evaluated within the same domain, limiting their generalizability.

\subsection{Zero-Shot Panorama Depth Estimation} Different from in-domain methods, zero-shot methods aim to perform robust predictions on unseen data. Depth AnyWhere \cite{wang2024depth} distills the knowledge from large perspective models \cite{yang2024depth} to small panorama models via unlabeled panorama data, while PanDA \cite{cao2025panda} distills from large panorama models. DA360 \cite{jiang2025depth} finetunes Depth Anything V2 \cite{yang2024depthv2} and equips the DPT \cite{ranftl2021vision} decoder with ERP padding. DA$^2$ \cite{li20252} generates high-quality panoramic depth data from perspective data via perspective-to-panorama diffusion \cite{tencent2025hunyuanworld}. DAP \cite{lin2026depth} introduces a three-stage curation pipeline to generate reliable pseudo labels for unlabeled images. DepthMaster \cite{wang2026depthmaster} and PaGeR \cite{bozic2026unified} utilize a Depth Anything V3 \cite{lin2025depth} structure to process cubemap images and produce ERP depth images. However, these zero-shot methods with large backbones tend to be computationally expensive, making them unsuitable for deployment on mobile platforms. Thus, we aim to develop a lightweight model by distilling complementary knowledge from panoramic and perspective teachers adaptively and enabling efficient spherical global context modeling.

\subsection{Zero-Shot Perspective Depth Estimation} Recent advances in monocular depth estimation from perspective
images have benefited from improved architectures and increasingly diverse training data. MiDaS \cite{ranftl2020towards} enables joint learning from heterogeneous datasets using a robust, scale-shift-invariant loss. Depth Anything \cite{yang2024depth} and Depth Anything V2 \cite{yang2024depthv2} leverage large-scale unlabeled images to improve generalization. Beyond relative depth estimation, MoGe2 \cite{wang2026moge} recovers metric-scale 3D point maps from single images using a decoupled design that separates relative geometry and metric-scale learning. Despite their strong generalization capabilities, their computation costs make them unfriendly to mobile platforms. To improve deployment efficiency, FastDepth \cite{wofk2019fastdepth} combines a compact encoder-decoder architecture with network pruning for embedded inference. More recently, ZipDepth \cite{tosi2026zipdepth} enables efficient zero-shot depth estimation with large-scale knowledge distillation from a depth foundation model. However, these approaches are designed for perspective images and cannot be applied directly to panorama images.

% Drop-in Method section. See main.tex for packages and references.bib for citations.
% Figures are relative to the compilation directory.
\section{Method}
\label{sec:method}

\subsection{Overview and our Baseline}
\label{sec:overview}

Given an equirectangular projection (ERP) image
$I\in\mathbb{R}^{H\times W\times 3}$, we estimate its metric depth map
$\hat{D}\in\mathbb{R}^{H\times W\times 1}$,
where $H$ and $W$ denote the image height and width.
The network predicts the inverse-depth map $\hat{s}$, and we recover metric depth as $\hat{D}=1/\hat{s}$.

We first finetune the base variant of ZipDepth \cite{tosi2026zipdepth} as our baseline since it has shown great accuracy-efficiency trade-offs in perspective depth estimation. Specifically, we initialize the model from the released official ZipDepth checkpoint and optimize metric inverse-depth objectives on labeled panorama data. Building on this baseline, we introduce APPD and SCFF to enhance its generalization to unseen data, resulting in our new model, EG360.

\subsection{Training Data and Complementary Teachers}
\label{sec:teachers}

\paragraph{Labeled and unlabeled panorama data}
Similar to PanDA \cite{cao2025panda}, we train our model on joint labeled synthetic data and unlabeled real-world data. Table \ref{tab:training_sources} reports the actual samples used.
The labeled set contains 562,815 RGB-depth pairs from
Structured3D \cite{zheng2020structured3d}, PanoCARLA \cite{songpvdepth}, and TartanAirV2 \cite{wang2020tartanair}.
Note that Structured3D RGB images rendered from the same viewpoint under three lighting conditions share the same ground-truth depth map, and TartanAir panoramas are rectified to an upright orientation offline following DA360 \cite{jiang2025depth}. The unlabeled set contains 73,701 real panoramas from ZInD \cite{cruz2021zillow} and HoliCity \cite{zhou2020holicity}.

\begin{table}[t]
\centering
\small
\setlength{\tabcolsep}{3pt}
\renewcommand{\arraystretch}{1.12}
\begin{tabular*}{\columnwidth}{
    @{\extracolsep{\fill}}lccr@{}
}
\toprule
Datasets & Indoor & Outdoor & Samples \\
\midrule
\multicolumn{4}{c}{\textit{Labeled Synthetic Datasets}} \\
\midrule
Structured3D \cite{zheng2020structured3d}
    & \ensuremath{\checkmark} & & 18,222 \\
PanoCARLA \cite{songpvdepth}
    & & \ensuremath{\checkmark} & 112,212 \\
TartanAirV2 \cite{wang2020tartanair}
    & \ensuremath{\checkmark}
    & \ensuremath{\checkmark} & 432,381 \\
\multicolumn{3}{l}{\textbf{Labeled total}}
    & \textbf{562,815} \\
\midrule
\multicolumn{4}{c}{\textit{Unlabeled Real-world Datasets}} \\
\midrule
ZInD \cite{cruz2021zillow}
    & \ensuremath{\checkmark} & & 67,448 \\
HoliCity \cite{zhou2020holicity}
    & & \ensuremath{\checkmark} & 6,253 \\
\multicolumn{3}{l}{\textbf{Unlabeled total}}
    & \textbf{73,701} \\
\bottomrule
\end{tabular*}
\caption{Labeled synthetic and unlabeled real-world datasets
used.}
\label{tab:training_sources}
\end{table}

\paragraph{Pseudo labels from teachers}
DAP operates on the ERP image, while MoGe2 processes six
$90^{\circ}$ cubemap faces and we use superscripts $\mathrm P$ and
$\mathrm C$ for the DAP and MoGe2 teachers, respectively.
Both teachers are frozen, and their predictions are cached offline. In particular, for each cubemap face $c \in \{1,\ldots,6\}$, MoGe2 predicts an axial depth map $z_c$,
and we convert it into radial depth map $D_c(u,v)$:
\begin{equation}
\begin{aligned}
D_c(u,v)
&= z_c(u,v)
\sqrt{
1+\left(\frac{u-c_x}{f}\right)^2
 +\left(\frac{v-c_y}{f}\right)^2
}
\end{aligned}
\label{eq:radial_conversion}
\end{equation}
Here, each cube face has a resolution of $R \times R$
pixels, and $(u,v)$ denotes the horizontal
and vertical pixel coordinates.
The principal point $(c_x,c_y)$ lies at the image center.
The focal length $f=R/2$ follows from the
$90^\circ$ field of view of each cube face.
The teacher targets $t^{\mathrm P}$ and $t^{\mathrm C}$ are inverse
radial depths, with $t_c^{\mathrm C}=1/D_c$.
Additionally, both teachers have a segmentation head that identifies the sky region in an outdoor image, and we set the distance of pixels in these regions to be 1000m. For labeled data, we also cap the distance to 1000m as pixels farther than 1000m are sky pixels.

\begin{table}[t]
\centering
\small
\setlength{\tabcolsep}{3pt}
\begin{tabular*}{\columnwidth}{@{\extracolsep{\fill}}lcccc@{}}
\toprule
& \multicolumn{2}{c}{Stanford2D3D}
& \multicolumn{2}{c}{Metropolis} \\
\cmidrule(lr){2-3}\cmidrule(lr){4-5}
Teacher
& AbsRel$\downarrow$ & $\delta_1\uparrow$
& AbsRel$\downarrow$ & $\delta_1\uparrow$ \\
\midrule
DAP \cite{lin2026depth}
& \textbf{0.071} & \textbf{0.965}
& 0.185 & 0.733 \\
MoGe2 \cite{wang2026moge}
& 0.150 & 0.774
& \textbf{0.127} & \textbf{0.862} \\
\bottomrule
\end{tabular*}
\caption{Comparison of panorama and perspective teachers. Metrics are averaged equally over six
cube faces for Stanford2D3D and four horizontal faces for
Metropolis, as its top and bottom faces lack valid
ground-truth depth. Best results are bolded.}
\label{tab:teacher_comparison}
\end{table}

\paragraph{Panorama and perspective teachers are complementary}
Different from previous perspective-teacher-only \cite{wang2024depth} or panorama-teacher-only \cite{cao2025panda} distillation, we design a joint
panorama-perspective distillation mechanism to combine the power from both worlds. To this end, we first verify that they are complementary across indoor and outdoor scenarios. Concretely, we evaluate the performance of the panoramic teacher DAP \cite{lin2026depth} and the
perspective teacher MoGe2 \cite{wang2026moge} on Stanford2D3D \cite{armeni2017joint} and Metropolis \cite{mapillary2024metropolis}, which are purely indoor and outdoor. For DAP, we input an ERP image, obtain its ERP depth, and convert it to CP depth. For MoGe2, we input CP images and get CP depth. Then evaluations are performed in the CP space in a face-wise manner. As shown in Table \ref{tab:teacher_comparison}, DAP performs better than MoGe2 on Stanford2D3D, whereas MoGe2 is more competitive on Metropolis. These observations motivate us to combine both panorama and perspective teachers for more effective distillation.

\subsection{Geometry-Guided Adaptive Teacher Weighting}
\label{sec:plane_weighting}

As shown in Table \ref{tab:teacher_comparison}, panorama and perspective teachers can be complementary, but the relative usefulness of the two teachers may vary across cube faces within the same panorama. A perspective view dominated by a wall may provide limited cues for estimating its absolute distance,
while the panoramic teacher can exploit the surrounding
scene. Thus, we use the predicted geometry of each cube
face to weigh the two teachers' supervision dynamically.

\begin{figure}[t]
    \centering
    \includegraphics[width=\linewidth]{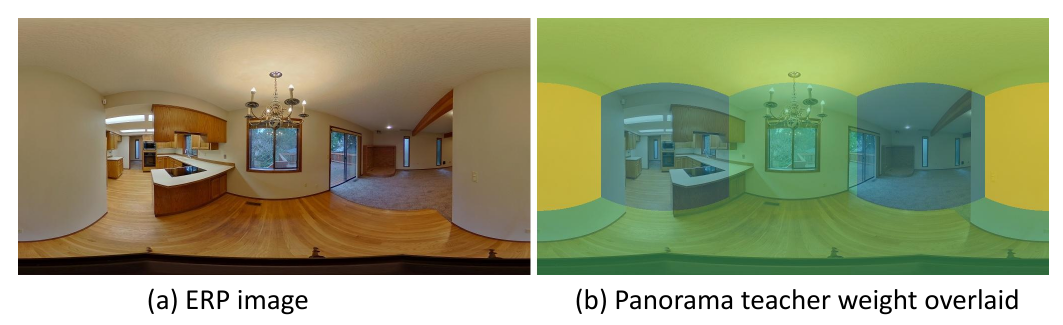}
    \caption{Visualization of geometry-adaptive teacher weighting. Faces dominated by planar surfaces, such as walls or ceilings, receive higher panorama teacher weights, while views with greater geometric variation receive lower weights (blue or green).}
    \label{fig:teacher_weight_overlay}
\end{figure}

\paragraph{Normalized plane-fitting residual}
For each valid, non-sky pixel $i$ on face $c$, we
reconstruct MoGe2 3D point as
$\mathbf{p}_{c,i}=\boldsymbol{\nu}_{c,i}/t^{\mathrm C}_{c,i}$,
where $t^{\mathrm C}_{c,i}$ is the projected inverse
radial depth and $\boldsymbol{\nu}_{c,i}$ is the
unit viewing ray. Then, we fit a single plane $\Pi_c$ to these points using PCA,
minimizing the sum of squared perpendicular distances to the plane.
Let $\mathcal{V}_c$ denote the set of valid pixels on face $c$.
We compute the mean radial depth $\bar{D}_c$ and the
normalized plane-fitting residual $r_c$ as:
\begin{equation}
\begin{aligned}
\bar{D}_c
&=\frac{1}{|\mathcal{V}_c|}
  \sum_{i\in\mathcal{V}_c}\|\mathbf{p}_{c,i}\|_2,\\
r_c
&=\frac{1}{\bar{D}_c}
  \sqrt{\frac{1}{|\mathcal{V}_c|}
  \sum_{i\in\mathcal{V}_c}
  \operatorname{dist}(\mathbf{p}_{c,i},\Pi_c)^2},
\end{aligned}
\label{eq:plane_residual}
\end{equation}
Here, $\bar{D}_c$ is the mean distance from the camera to these points,
and $\operatorname{dist}$ denotes the distance to the fitted plane. Intuitively, a smaller $r_c$ indicates more planar geometry.

\paragraph{Teacher weights}
We assign complementary weights to the panoramic
teacher DAP ($\mathrm P$) and the perspective teacher
MoGe2 ($\mathrm C$):
\begin{equation}
w_c^{\mathrm P}=\frac{\tau}{r_c+\tau},
\qquad
w_c^{\mathrm C}=\frac{r_c}{r_c+\tau},
\qquad \tau>0.
\label{eq:teacher_weights}
\end{equation}
Thus, more planar faces receive stronger panorama teacher supervision,
while larger residuals increase the perspective teacher's contribution.
The parameter $\tau$ controls this balance, with equal
weights at $r_c=\tau$. In practice, we set $\tau$ to be 0.3 experimentally. As shown in Fig. \ref{fig:teacher_weight_overlay}, our weighting is adaptive based on the view geometry.

\paragraph{Weighted L1 distillation}
We combine DAP supervision in ERP space and MoGe2
supervision in cubemap space.
For one panorama, the distillation loss is
\begin{equation}
\begin{aligned}
\mathcal{L}_{\mathrm{distill}}
={}&\frac{1}{N_{\mathrm P}}
\sum_{i=1}^{N_{\mathrm P}}
w_i^{\mathrm P}|\hat{s}_i-t_i^{\mathrm P}|\\
&+\frac{1}{|\mathcal{F}|}
\sum_{c\in\mathcal{F}}\frac{w_c^{\mathrm C}}{N_c}
\sum_{j=1}^{N_c}
|\hat{s}_{c,j}-t_{c,j}^{\mathrm C}|,
\end{aligned}
\label{eq:joint_distillation}
\end{equation}
Here, $\hat{s}$ and $t$ denote student and teacher inverse depths.
The value $\hat{s}_{c,j}$ is obtained by projecting the student's
ERP inverse-depth prediction onto cube face $c$. $N_{\mathrm P}$ is the number of valid pixels for the panorama teacher and $N_c$ is the number of valid pixels for the perspective teacher on cube $c$.
The set $\mathcal{F}$ contains valid cube faces.
Each ERP pixel receives the DAP weight of its corresponding cube face.

\subsection{Spherical Cross-Level Feature Fusion}
\label{sec:spherical_refinement}

\begin{figure}[t]
    \centering
    \includegraphics[width=0.5\textwidth]{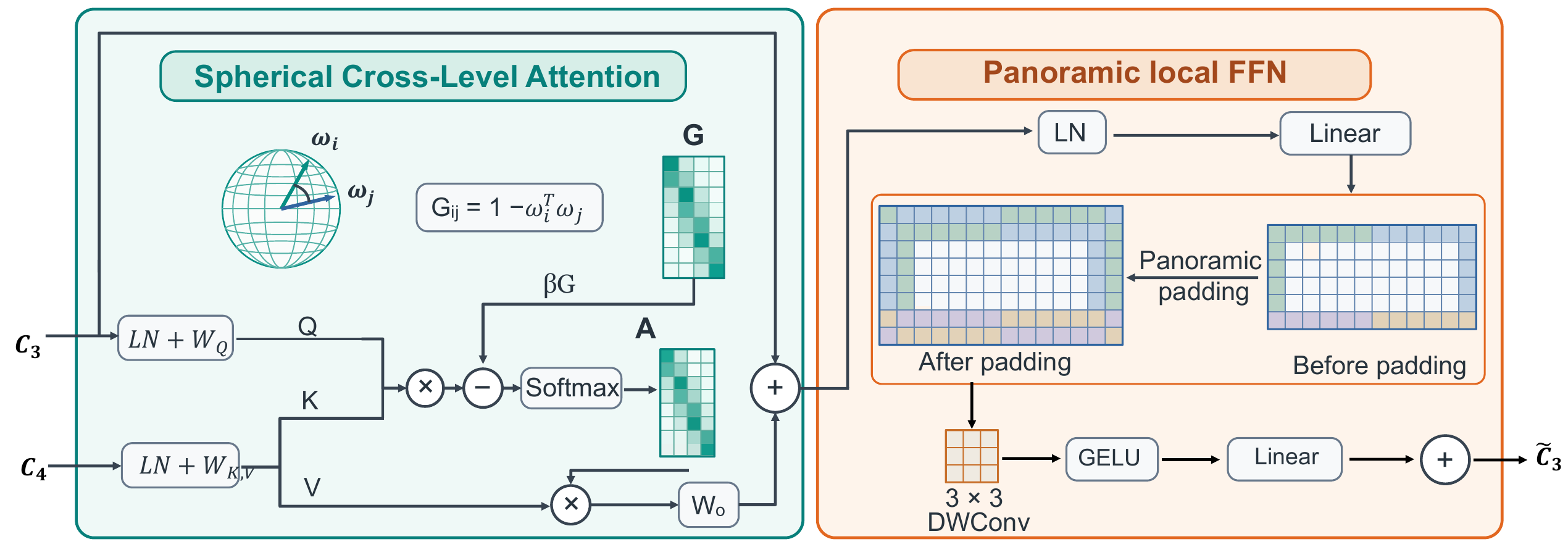}
    \caption{Diagram of Spherical Cross-Level Feature Fusion. Cross-attention with a spherical direction bias is followed by a local feed-forward network with panorama padding and depthwise convolution. Reshaping is omitted for brevity.}
    \label{fig:spherical_refinement}
\end{figure}

Previous works have shown that perceiving the global context is important for depth estimation \cite{zhang2023lite,ning2023trap}. Also, our ZipDepth baseline already implements several modules to enhance global perception. However, due to the distortion and left-right discontinuity in ERP images, its global modeling is insufficient, so we design a Spherical Cross-Level Feature Fusion (SCFF) module to further enhance its contextual modeling capabilities. As shown in Fig. \ref{fig:spherical_refinement}, it consists of a spherical cross-level attention module and a panorama local feed-forward network. 

\paragraph{Cross-level attention}
Let $C_3$ and $C_4$ denote the outputs of the third
and fourth encoder stages.
We flatten their spatial dimensions into token
sequences $X_3$ and $X_4$, respectively.
Queries, keys, and values are computed as
$Q=\operatorname{LN}(X_3)W_Q$,
$K=\operatorname{LN}(X_4)W_K$, and
$V=\operatorname{LN}(X_4)W_V$.
Here, $\operatorname{LN}$ denotes layer normalization \cite{ba2016layernormalization},
and $W_Q$, $W_K$, and $W_V$ are learnable projection
matrices. Bias terms are omitted for brevity.
In this way, each fine-scale query can therefore aggregate
information from the coarse feature.

\paragraph{Spherical direction bias}
Furthermore, we incorporate spherical geometry into attention
to account for the spatial relationships represented
by ERP images.
For a feature token at normalized pixel-center
coordinates $(x,y)$, its longitude and latitude are
$\lambda=2\pi(x-\frac12)$ and
$\phi=\pi(\frac12-y)$.
Its corresponding unit viewing direction is
\begin{equation}
\boldsymbol{\omega}
=
[\cos\phi\cos\lambda,\ \sin\phi,\ \cos\phi\sin\lambda]
^{\mathsf T}.
\label{eq:spherical_directions}
\end{equation}
We compute these directions for feature grids
and construct a distance matrix $\mathbf{G}$, where
$G_{ij}=1-\boldsymbol{\omega}_i^{\mathsf T}
\boldsymbol{\omega}_j$ measures the directional
distance between query $i$ and key $j$.
After splitting $Q$, $K$, and $V$ into heads, each head computes
(head indices are omitted for brevity)
\begin{equation}
A=\operatorname{softmax}
\left(
\frac{QK^{\mathsf T}}{\sqrt{d_h}}
-\beta\mathbf{G}
\right),
\label{eq:spherical_attention}
\end{equation}
Here, $d_h$ is the query/key dimension of each head.
Each head has its own positive bias strength
$\beta=\operatorname{softplus}(b)$, with learnable
parameter $b$.
This spherical direction bias favors nearby spherical directions,
including neighbors across the ERP left-right seam.
Let $Z$ denote the outputs $AV$ from all heads.
The residual update is $U=X_3+ZW_O$, where the learnable projection
$W_O$ maps $Z$ back to the channel dimension of $X_3$.

\paragraph{Panoramic local feed-forward network.}
We insert a local panoramic feed-forward network after the attention:
\begin{equation}
\widetilde{X}_3
=
U+
\operatorname{GELU}
\left(
\operatorname{DWConv}^{\mathrm{ERP}}_{3\times3}
\left(\operatorname{LN}(U)W_1\right)
\right)W_2.
\label{eq:panoramic_ffn}
\end{equation}
The matrices $W_1$ and $W_2$ are the learnable input and output
projections of the feed-forward network, respectively.
The $\operatorname{DWConv}^{\mathrm{ERP}}_{3\times3}$ is a depthwise convolution with panoramic padding \cite{jiang2025depth}. Following the zero-initialization strategy of
ControlNet~\cite{zhang2023adding}, we initialize
the output projections $W_O$ and $W_2$ to zero.
In the end, we reshape $\widetilde{X}_3$ to refined $\widetilde{C}_3$, which is passed together with $C_4$ to the original decoder.

\subsection{Overall loss function}
\label{sec:objectives}

\paragraph{Labeled data loss}
For labeled data, $D^{*}$ denotes the ground-truth metric depth map
and $s^{*}=1/D^{*}$ its inverse-depth map.
Let $\mathcal{M}$ contain all valid pixels in the supervised batch.
At pixel $p$, $\hat{s}_p$ and $s^{*}_p$ denote the predicted
and ground-truth inverse depths, respectively.
The supervised loss combines metric inverse-depth regression
with a multiscale gradient term \cite{jiang2025depth}:
\begin{equation}
 \mathcal{L}_{\mathrm{sup}}=
 \frac{1}{|\mathcal{M}|}\sum_{p\in\mathcal{M}}
 |\hat{s}_p-s^{*}_p|
 +\lambda_{\mathrm g}\mathcal{L}_{\mathrm{grad}}
 \label{eq:supervised_loss}
\end{equation}
The $\lambda_{\mathrm g}$ controls the gradient loss weight and is set to 1.

\paragraph{Overall objective}
For training, the overall objective combines the supervised loss and the distillation loss:
\begin{equation}
 \mathcal{L}_{\mathrm{total}}
 =\mathcal{L}_{\mathrm{sup}}+\mathcal{L}_{\mathrm{distill}}.
 \label{eq:total_loss}
\end{equation}

% Verified on 2026-11-05 against per-dataset result.json files and saved args.json.
% Local provenance: experiment_audit/verified_data.json (paths and SHA256 hashes).
% EG360 and APPD ablation rows: CP T=0.3, verified in experiment_audit/table3_cp03_verified.json.
% Main DA360-S: supervised fine-tune, division1000, 2026-10-18_14-33-33.
% Structural Full: original DLC18 e08, 2026-10-27_14-58-08; not the later e08_full repeat.
% Efficiency updated on 2026-11-09 from logs/orin_benchmark_20261109/formal_run/formal/*.json.
% All ZipDepth variants use their _reparam results; EG360 maps to zipdepth_full_reparam.
% Latency: 300 timed forwards, random initialization; accuracy above retains trained checkpoints.
% DepthMaster/PaGeR failed; their Params/GFLOPs retain audited architectural reference values.
% APPD transfer: CP tau=0.3, final step 100000; verified in experiment_audit/transfer_cp03_verified_20261107.json.
\section{Experiments}
\label{sec:experiments}

\subsection{Experimental Setup}
\label{sec:exp_setup}

\begin{table*}[t]
\centering
\footnotesize
\setlength{\tabcolsep}{3pt}
\renewcommand{\arraystretch}{1.12}
\resizebox{0.95\textwidth}{!}{%
\begin{tabular}{@{}lrrrrrrrrr@{}}
\toprule
& & & & \multicolumn{2}{c}{Matterport3D} & \multicolumn{2}{c}{Stanford2D3D} & \multicolumn{2}{c}{Metropolis} \\
\cmidrule(lr){5-6}\cmidrule(lr){7-8}\cmidrule(lr){9-10}
Method & Params (M) & GFLOPs$\downarrow$ & Latency (ms)$\downarrow$ & AbsRel$\downarrow$ & $\delta_1\uparrow$ & AbsRel$\downarrow$ & $\delta_1\uparrow$ & AbsRel$\downarrow$ & $\delta_1\uparrow$ \\
\midrule
\multicolumn{10}{l}{\textit{Foundation models}} \\ 
\midrule
DA360-S$^{\dagger}$ \cite{jiang2025depth} & \textbf{24.804} & \textbf{299.520} & \textbf{517.412} & 0.168 & 0.802 & 0.082 & 0.947 & 0.232 & 0.597 \\
DAP \cite{lin2026depth} & 365.050 & 2543.697 & 1868.322 & 0.137 & 0.840 & 0.072 & 0.963 & \textbf{0.195} & \textbf{0.734} \\
DepthMaster \cite{wang2026depthmaster} & 385.191 & 11521.707 & OOM & \textbf{0.101} & \textbf{0.905} & \textbf{0.062} & 0.965 & 1.022 & 0.263 \\
PaGeR \cite{bozic2026unified} & 1286.563 & 26200.188 & OOM & 0.116 & 0.889 & 0.064 & \textbf{0.972} & 0.224 & 0.500 \\
\midrule 
\multicolumn{10}{l}{\textit{Lightweight models}} \\
\midrule
UniFuse$^{\dagger}$ \cite{jiang2021unifuse} & 30.263 & 124.592 & 90.863 & 0.234 & 0.663 & 0.155 & 0.754 & 0.297 & 0.500 \\
BiFuse++$^{\dagger}$ \cite{wang2022bifuse++} & 32.272 & 106.598 & 71.284 & 0.295 & 0.555 & 0.204 & 0.717 & 0.398 & 0.260 \\
ACDNet$^{\dagger}$ \cite{zhuang2022acdnet} & 46.606 & 156.655 & 140.114 & 0.233 & 0.670 & 0.146 & 0.795 & 0.301 & 0.487 \\
EG360(ours) & \textbf{6.593} & \textbf{23.492} & \textbf{28.509} & \textbf{0.173} & \textbf{0.784} & \textbf{0.094} & \textbf{0.923} & \textbf{0.237} & \textbf{0.632} \\
\bottomrule
\end{tabular}}
\caption{Comparison with existing method without ground-truth scale alignment. $\dagger$ denotes models retrained or finetuned with our recipe. Latency is measured on a Jetson Orin NX 16GB in MAXN mode, with AMP FP16 and batch size 1. EG360 uses structural reparameterization. OOM denotes out of memory. Bold denotes the best value in each column within each group.}
\label{tab:exp_main}
\end{table*}

\paragraph{Data for training and evaluation}
We use the labeled synthetic and unlabeled real panoramas in Table~\ref{tab:training_sources} for training. We evaluate cross-dataset generalization on the Matterport3D \cite{chang2017matterport3d}, Stanford2D3D \cite{armeni2017joint}, and Metropolis \cite{mapillary2024metropolis} test sets, containing 2,014, 373, and 3,000 panoramas, respectively, in a zero-shot manner. The maximum evaluation depth is 10 meters for Matterport3D and Stanford2D3D, and 100 meters for Metropolis. We report the absolute relative error (AbsRel) and threshold accuracy ($\delta_1$). For each image, let $\mathcal{V}$ denote the set of valid evaluation pixels, $|\mathcal{V}|$ its size, and $\hat d_p$ and $d_p^*$ the predicted and ground-truth depths at pixel $p$, respectively. AbsRel is the mean absolute depth error normalized by the ground-truth depth, $\mathrm{AbsRel}=\frac{1}{|\mathcal{V}|}\sum_{p\in\mathcal{V}}\frac{|\hat d_p-d_p^*|}{d_p^*}$. The threshold accuracy $\delta_1$ is the fraction of pixels $p\in\mathcal{V}$ satisfying $\max(\hat d_p/d_p^*,d_p^*/\hat d_p)<1.25$. Note that all evaluations are done without any alignment.

\paragraph{Implementation Details}
We initialize the base student from the released ZipDepth checkpoint and train each of our configurations independently for 100,000 iterations at $512\times1024$ ERP resolution using the PyTorch \cite{paszke2019pytorchimperativestylehighperformance} framework on an RTX 2080 Ti. We use AdamW \cite{loshchilov2017decoupled} with weight decay $10^{-3}$, gradient-norm clipping at 1.000, and AMP FP16. We also use a OneCycle \cite{smith2019super} learning schedule with a peak learning rate of $10^{-4}$, a 1\% warm-up fraction, and linear annealing. Unless stated otherwise, in each iteration we use a batch size of 8 for labeled data and 8 for unlabeled data when using distillation. We use the checkpoint from the last step to evaluate for fairness. Augmentation includes horizontal flips, color jitter, random yaw and pitch rotations, and polar RGB masking, following DA360 \cite{jiang2025depth}.

\paragraph{Comparison settings}
For UniFuse, BiFuse++, and ACDNet, we retrain them using ImageNet-pretrained ResNet-18 encoders. For DA360-S, the ViT-Small version, we finetune its released checkpoint. These four models are retrained on our labeled data for fairness. For DAP, DepthMaster, and PaGeR, we directly use released weights. Efficiency is measured on Jetson Orin NX 16GB (MAXN, PyTorch eager, AMP FP16, batch size one, native input resolutions). Latency averages 300 forward passes after warm-up, excluding I/O and CPU preprocessing. Attention uses native PyTorch Scaled Dot-Product Attention (SDPA) with external xFormers \cite{xFormers2022} and flashAttn \cite{dao2022flashattention, dao2023flashattention2} disabled for fairness and pytorch-native. Our model uses reparameterization \cite{ding2021repvgg}. DepthMaster and PaGeR did not complete this formal run due to OOM.

\subsection{Comparison with Existing Methods}
\label{sec:exp_comparison}

Table \ref{tab:exp_main} compares depth accuracy and inference cost, with latency measured on a Jetson Orin NX 16GB in MAXN mode. Compared with UniFuse, EG360 reduces AbsRel by 26.245\%, 39.432\%, and 20.297\% on Matterport3D, Stanford2D3D, and Metropolis, respectively, and also improves $\delta_1$ on all three datasets.
With only 6.593M parameters and 23.492 GFLOPs, EG360 is also more efficient than previous lightweight models.
Compared to DA360-S, EG360 achieves similar results on Matterport3D and Metropolis while running 18 times faster. DAP and PaGeR can achieve better results, but they require 2543.697 and 26200.188 GFLOPs of computation. As for DepthMaster, it achieves the lowest indoor AbsRel, but its high Metropolis error may be related to the absence of outdoor panoramas in training.

\begin{figure*}[t]
\centering
\includegraphics[width=0.95\textwidth]{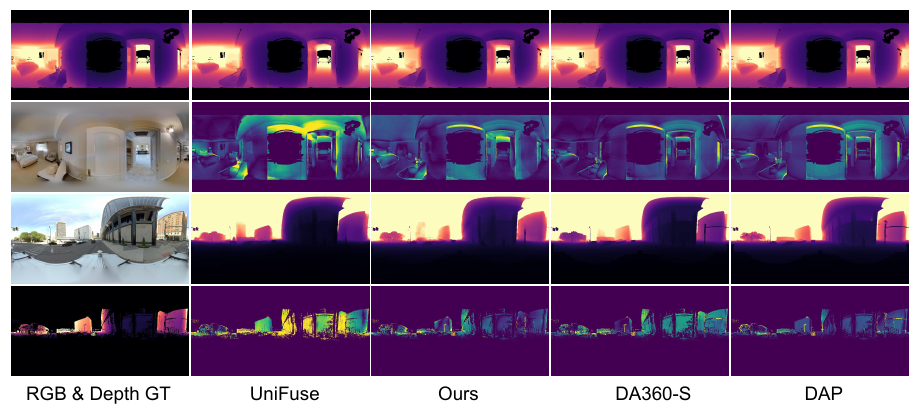}
\caption{Qualitative comparisons on Matterport3D (top) and Metropolis (bottom). The first column shows ground-truth depth and RGB images; the remaining columns show predictions above per-pixel AbsRel maps for UniFuse, EG360, DA360-S, and DAP. Error colors range from purple (low) to yellow (high). Pixels without ground truth depth are masked in the error map.}
\label{fig:exp_qualitative_comparison}
\end{figure*}

\paragraph{Qualitative comparisons}
Figure \ref{fig:exp_qualitative_comparison} shows qualitative results from indoor and outdoor scenarios. For example, our EG360 can obtain lower errors at the distorted door in the indoor scene and the far buildings in the outdoor scene than UniFuse. Additionally, EG360 can achieve similar results against DA360-S and DAP, as shown by the depth prediction and error maps.

\subsection{Ablation Studies}
\label{sec:exp_overall}

\begin{table}[t]
\centering
\scriptsize
\setlength{\tabcolsep}{2pt}
\renewcommand{\arraystretch}{1.12}
\resizebox{\columnwidth}{!}{%
\begin{tabular}{@{}lrrrrrrrr@{}}
\toprule
& & & \multicolumn{2}{c}{Matterport3D} & \multicolumn{2}{c}{Stanford2D3D} & \multicolumn{2}{c}{Metropolis} \\
\cmidrule(lr){4-5}\cmidrule(lr){6-7}\cmidrule(lr){8-9}
Method & Params (M) & GFLOPs$\downarrow$ & AbsRel$\downarrow$ & $\delta_1\uparrow$ & AbsRel$\downarrow$ & $\delta_1\uparrow$ & AbsRel$\downarrow$ & $\delta_1\uparrow$ \\
\midrule
Baseline & \textbf{6.144} & \textbf{21.196} & 0.199 & 0.722 & 0.126 & 0.850 & 0.266 & 0.578 \\
+ APPD & \textbf{6.144} & \textbf{21.196} & \textbf{0.169} & 0.781 & 0.101 & 0.903 & 0.244 & 0.619 \\
+ SCFF & 6.593 & 23.492 & 0.195 & 0.733 & 0.113 & 0.876 & 0.263 & 0.582 \\
EG360 & 6.593 & 23.492 & 0.173 & \textbf{0.784} & \textbf{0.094} & \textbf{0.923} & \textbf{0.237} & \textbf{0.632} \\
\bottomrule
\end{tabular}}
\caption{Overall ablation of APPD and SCFF on three real-world test datasets. EG360 includes both components. Bold denotes the best value in each column.}
\label{tab:exp_overall}
\end{table}

\paragraph{Overall Abaltion Study} Table \ref{tab:exp_overall} report the overall ablation study for APPD and SCFF. APPD alone reduces AbsRel from 0.199 to 0.169 on Matterport3D, from 0.126 to 0.101 on Stanford2D3D, and from 0.266 to 0.244 on Metropolis. SCFF alone also improves all six metrics over the baseline without using pseudo labels. Combining the two components yields EG360, which achieves the best overall results among test datasets. Especially, SCFF is more effective on Stanford2D3D, as it's intended for global semantic enhancement, which could be more useful for indoor scenarios. In terms of efficiency, APPD leaves the inference architecture unchanged at 6.144M parameters and 21.196 GFLOPs while SCFF adds 0.449M parameters, 2.296 GFLOPs.

\label{sec:exp_distillation}
\begin{table}[t]
\centering
\scriptsize
\setlength{\tabcolsep}{2.5pt}
\renewcommand{\arraystretch}{1.12}
\resizebox{\columnwidth}{!}{%
\begin{tabular}{@{}lrrrrrr@{}}
\toprule
& \multicolumn{2}{c}{Matterport3D} & \multicolumn{2}{c}{Stanford2D3D} & \multicolumn{2}{c}{Metropolis} \\
\cmidrule(lr){2-3}\cmidrule(lr){4-5}\cmidrule(lr){6-7}
Distill Method & AbsRel$\downarrow$ & $\delta_1\uparrow$ & AbsRel$\downarrow$ & $\delta_1\uparrow$ & AbsRel$\downarrow$ & $\delta_1\uparrow$ \\
\midrule
No distill & 0.199 & 0.722 & 0.126 & 0.850 & 0.266 & 0.578 \\
DAP only & 0.176 & 0.771 & \textbf{0.101} & \textbf{0.901} & 0.257 & 0.601 \\
MoGe2 only & 0.176 & 0.762 & 0.146 & 0.762 & \textbf{0.235} & 0.628 \\
DAP + MoGe2 (Avg) & \textbf{0.168} & \textbf{0.780} & 0.109 & 0.889 & 0.238 & \textbf{0.630} \\
\bottomrule
\end{tabular}}
\caption{Teacher complementarity with our baseline student. Using both teachers leads to better results.}
\label{tab:exp_teachers}
\end{table}

\paragraph{Ablation on joint and single panorama-perspective distillation}
As discussed in Table \ref{tab:teacher_comparison}, DAP and MoGe2 are complementary in terms of performance on different domains. Here, as listed in Table \ref{tab:exp_teachers}, simply averaging the losses from the two teachers yields more balanced results than using them alone. For example, on Matterport3D, simple averaging yields an AbsRel of 0.168 while using them alone leads to 0.176.

\begin{table}[t]
\centering
\scriptsize
\setlength{\tabcolsep}{2.5pt}
\renewcommand{\arraystretch}{1.12}
\resizebox{\columnwidth}{!}{%
\begin{tabular}{@{}lrrrrrr@{}}
\toprule
& \multicolumn{2}{c}{Matterport3D} & \multicolumn{2}{c}{Stanford2D3D} & \multicolumn{2}{c}{Metropolis} \\
\cmidrule(lr){2-3}\cmidrule(lr){4-5}\cmidrule(lr){6-7}
Batch size & AbsRel$\downarrow$ & $\delta_1\uparrow$ & AbsRel$\downarrow$ & $\delta_1\uparrow$ & AbsRel$\downarrow$ & $\delta_1\uparrow$ \\
\midrule
1 & 0.184 & 0.758 & 0.115 & 0.875 & 0.240 & 0.630 \\
2 & 0.175 & 0.769 & 0.114 & 0.872 & 0.242 & 0.613 \\
4 & 0.173 & 0.778 & 0.110 & 0.883 & 0.244 & 0.618 \\
8 & 0.168 & 0.780 & 0.109 & 0.889 & \textbf{0.238} & \textbf{0.630} \\
16 & \textbf{0.166} & \textbf{0.782} & \textbf{0.108} & \textbf{0.890} & 0.239 & 0.628 \\
\bottomrule
\end{tabular}}
\caption{Effect of the batch size for unlabeled data under Avg distillation. The first column is the unlabeled batch size.}
\label{tab:exp_batch}
\end{table}

\paragraph{Ablation of batch size of unlabeled data}
In Table ~\ref{tab:exp_batch}, we vary the unlabeled batch size from 1 to 16 under average distillation. Increasing it from 1 to 8 reduces Matterport3D AbsRel from 0.184 to 0.168 and Stanford2D3D AbsRel from 0.115 to 0.109. A batch size of 16 gives similar results against a batch size of 8 but would leads to more computation during training. Thus, we use 8 as the unlabeled batch size across our other experiments. 

\begin{table}[t]
\centering
\scriptsize
\setlength{\tabcolsep}{2.5pt}
\renewcommand{\arraystretch}{1.12}
\resizebox{\columnwidth}{!}{%
\begin{tabular}{@{}lrrrrrr@{}}
\toprule
& \multicolumn{2}{c}{Matterport3D} & \multicolumn{2}{c}{Stanford2D3D} & \multicolumn{2}{c}{Metropolis} \\
\cmidrule(lr){2-3}\cmidrule(lr){4-5}\cmidrule(lr){6-7}
Distill Method & AbsRel$\downarrow$ & $\delta_1\uparrow$ & AbsRel$\downarrow$ & $\delta_1\uparrow$ & AbsRel$\downarrow$ & $\delta_1\uparrow$ \\
\midrule
Avg & 0.168 & 0.780 & 0.109 & 0.889 & \textbf{0.238} & \textbf{0.630} \\
Random & 0.172 & 0.770 & 0.113 & 0.874 & 0.240 & 0.625 \\
Manual & \textbf{0.166} & 0.777 & 0.107 & 0.885 & 0.238 & 0.604 \\
APPD & 0.169 & \textbf{0.781} & \textbf{0.101} & \textbf{0.903} & 0.244 & 0.619 \\
\bottomrule
\end{tabular}}
\caption{Joint teacher strategies. Avg averages the two teacher losses. Random selects one teacher with equal probability per unlabeled batch. Manual assigns DAP to ZInD and MoGe2 to HoliCity. APPD is our geometry-adaptive loss weighting.}
\label{tab:exp_fusion}
\end{table}

\paragraph{Ablation on how to combine teachers}
In Table~\ref{tab:exp_fusion}, Avg averages both teacher losses at every iteration. Random \cite{he2025distilldepthdistillationcreates} selects one teacher with equal probability for the entire unlabeled batch. Manual assigns DAP to ZInD and MoGe2 to HoliCity. APPD is our geometry-guided distillation, which reduces Stanford2D3D AbsRel by 7.411\% relative to Avg and increases $\delta_1$ from 0.889 to 0.903. It also attains the highest Matterport3D $\delta_1$. That is, APPD offers a useful accuracy trade-off across various domains.

\begin{table}[t]
\centering
\scriptsize
\setlength{\tabcolsep}{2.5pt}
\renewcommand{\arraystretch}{1.12}
\resizebox{\columnwidth}{!}{%
\begin{tabular}{@{}lrrrrrr@{}}
\toprule
& \multicolumn{2}{c}{Matterport3D} & \multicolumn{2}{c}{Stanford2D3D} & \multicolumn{2}{c}{Metropolis} \\
\cmidrule(lr){2-3}\cmidrule(lr){4-5}\cmidrule(lr){6-7}
Loss & AbsRel$\downarrow$ & $\delta_1\uparrow$ & AbsRel$\downarrow$ & $\delta_1\uparrow$ & AbsRel$\downarrow$ & $\delta_1\uparrow$ \\
\midrule
L1 & \textbf{0.168} & \textbf{0.780} & \textbf{0.109} & 0.889 & \textbf{0.238} & \textbf{0.630} \\
SI & 0.228 & 0.680 & 0.155 & 0.805 & 0.246 & 0.603 \\
LSS & 0.176 & 0.748 & 0.111 & \textbf{0.901} & 0.466 & 0.095 \\
\bottomrule
\end{tabular}}
\caption{Pseudo-label loss type ablation under Avg teacher fusion.}
\label{tab:exp_pseudo_loss}
\end{table}

\paragraph{Ablation on loss type for pseudo labels}
Table \ref{tab:exp_pseudo_loss} compares three losses for pseudo labels while keeping both teachers fixed. SI normalizes each map by its dispersion as in DA360 \cite{jiang2025depth}, while LSS fits a scale by least squares before computing the L1 residual. L1 achieves the best result on five of the six metrics. We can see that with SI or LSS, the model would perform much worse on either indoor or outdoor datasets. This indicates that the training cannot recover the actual scale correctly, as these two losses are scale-invariant. Thus, we use L1 loss for unlabeled data.

\begin{table}[t]
\centering
\scriptsize
\setlength{\tabcolsep}{2.5pt}
\renewcommand{\arraystretch}{1.12}
\resizebox{\columnwidth}{!}{%
\begin{tabular}{@{}lrrrrrr@{}}
\toprule
& \multicolumn{2}{c}{Matterport3D} & \multicolumn{2}{c}{Stanford2D3D} & \multicolumn{2}{c}{Metropolis} \\
\cmidrule(lr){2-3}\cmidrule(lr){4-5}\cmidrule(lr){6-7}
Method & AbsRel$\downarrow$ & $\delta_1\uparrow$ & AbsRel$\downarrow$ & $\delta_1\uparrow$ & AbsRel$\downarrow$ & $\delta_1\uparrow$ \\
\midrule
UniFuse & 0.234 & 0.663 & 0.155 & 0.754 & 0.297 & 0.500 \\
+ APPD & \textbf{0.197} & \textbf{0.745} & \textbf{0.119} & \textbf{0.867} & \textbf{0.250} & \textbf{0.603} \\
\midrule
BiFuse++ & 0.295 & 0.555 & 0.204 & 0.717 & 0.398 & 0.260 \\
+ APPD & \textbf{0.211} & \textbf{0.718} & \textbf{0.158} & \textbf{0.760} & \textbf{0.270} & \textbf{0.546} \\
\midrule
ACDNet & 0.233 & 0.670 & 0.146 & 0.795 & 0.301 & 0.487 \\
+ APPD & \textbf{0.190} & \textbf{0.753} & \textbf{0.114} & \textbf{0.873} & \textbf{0.253} & \textbf{0.600} \\
\bottomrule
\end{tabular}}
\caption{Applying APPD to other lightweight models.}
\label{tab:exp_transfer}
\end{table}

\paragraph{Applying APPD to other architectures}
Table \ref{tab:exp_transfer} reports results of applying APPD to UniFuse, BiFuse++, and ACDNet.
Clearly, all six reported metrics have been improved. For BiFuse++, its Matterport3D AbsRel decreases from 0.295 to 0.211, and its Metropolis $\delta_1$ increases from 0.260 to 0.546.
These results show that APPD can be applied to other models, either using ERP images alone \cite{zhuang2022acdnet} or using both the ERP and CP images \cite{jiang2025depth}.

\begin{figure}[t]
\centering
\includegraphics[width=0.5\textwidth]{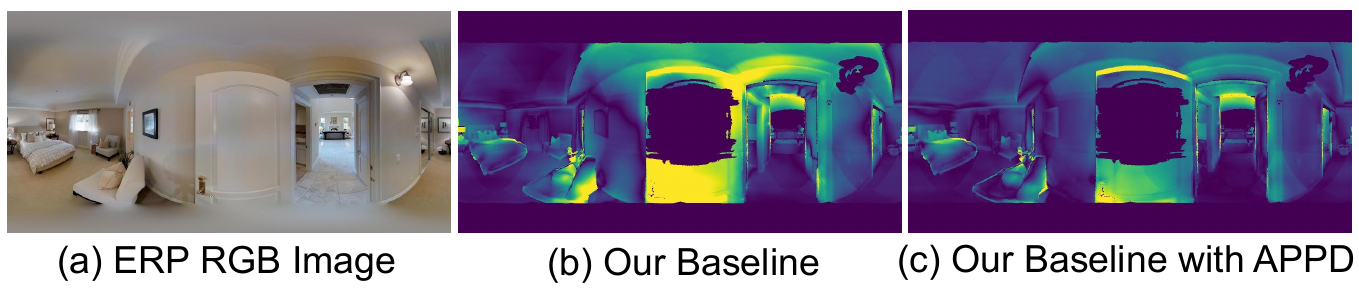}
\caption{Qualitative results of applying APPD to our baseline model. The error is reduced in this real-world example.}
\label{fig:exp_distillation_qualitative}
\end{figure}

\paragraph{Qualitative effect of APPD}
Figure \ref{fig:exp_distillation_qualitative} compares the results of applying APPD or not to our baseline. From the error maps, it is clear that APPD substantially reduces the high errors 
at the sofa and the door, which are challenging.

\paragraph{Ablation on components of SCFF}
\label{sec:exp_refinement}

\begin{figure}[t]
\centering
\includegraphics[width=0.5\textwidth]{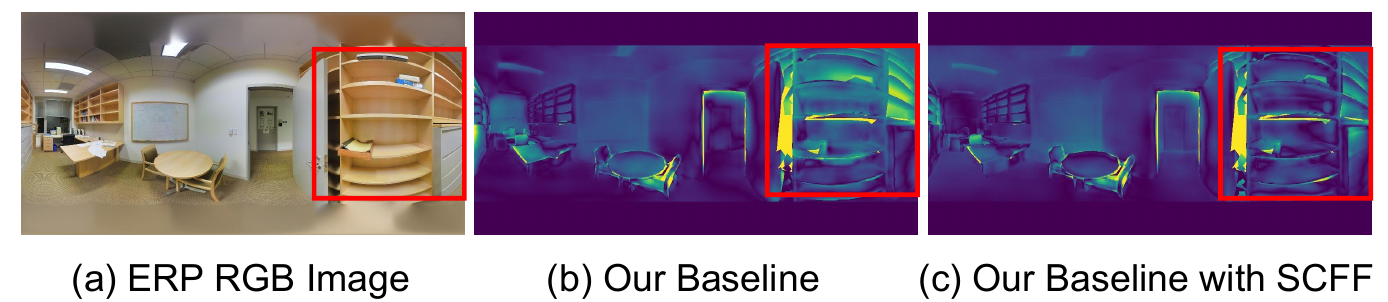}
\caption{Qualitative results of using SCFF. The error for the large shelf is lowered as SCFF allows long-range context modeling.}
\label{fig:exp_cross_level_qualitative}
\end{figure}

\begin{table}[t]
\centering
\small
\setlength{\tabcolsep}{3pt}
\renewcommand{\arraystretch}{1.12}
\resizebox{\columnwidth}{!}{%
\begin{tabular}{@{}lrrrr@{}}
\toprule
Configuration & Params (M) & GFLOPs$\downarrow$ & AbsRel$\downarrow$ & $\delta_1\uparrow$ \\
\midrule
Baseline & \textbf{6.144} & \textbf{21.196} & 0.126 & 0.850 \\
Full & 6.593 & 23.492 & \textbf{0.113} & \textbf{0.876} \\
Full $-$ D & 6.593 & 23.492 & 0.125 & 0.864 \\
Full $-$ Cross & 6.296 & 21.814 & 0.119 & 0.867 \\
Full $-$ $3\!\times\!3$ DW Conv & 6.589 & 23.478 & 0.122 & 0.868 \\
Full $-$ Cross $+$ C4 Self & 6.839 & 22.173 & 0.119 & 0.875 \\
Full $-$ Cross $+$ C3 Self & 6.494 & 26.914 & 0.118 & 0.874 \\
\bottomrule
\end{tabular}}
\caption{SCFF ablation on Stanford2D3D using labeled data only.
Full is the complete SCFF module. $-$ D removes the spherical direction bias, $-$ Cross removes cross-level attention, and $-$ $3\times3$ DW Conv removes the depthwise convolution in the panoramic FFN.
Full $-$ Cross $+$ C4 Self/C3 Self replaces the cross attention with a self-attention block on C4/C3.}
\label{tab:exp_structure}
\end{table}

Table~\ref{tab:exp_structure} evaluates SCFF on Stanford2D3D under supervised training, without pseudo labels.
Full reduces AbsRel from 0.126 to 0.113 and increases $\delta_1$ from 0.850 to 0.876, adding 2.296 GFLOPs.
Removing direction bias D, cross-level attention, or the panoramic $3\times3$ depthwise convolution degrades accuracy.
Self-attention on C4 or C3 also yields lower accuracy than our cross-level attention.

\paragraph{Qualitative effect of using SCFF}
Figure \ref{fig:exp_cross_level_qualitative} shows reduced errors over the highlighted shelf surfaces after adding SCFF. This example thus depicts the benefit of global feature fusion of the model with spherical geometry injected.

\section{Conclusion}
\label{sec:conclusion}

In this work, we presented EG360, an efficient and generalizable panorama depth estimation model built on top of our Adaptive
Panorama–Perspective Distillation technique and Spherical Cross-Level Feature Fusion module. Experiments demonstrate that our model can achieve strong generalization ability while enabling real-time inference on a Jetson Orin NX 16GB in MAXN mode.

Limitations and future work. Although EG360 can achieve competitive results, we only used limited unlabeled data, and scaling the unlabeled data may further enhance performance \cite{xue2026depthart, wallingford2024image}. Additionally, as a single-frame model, it suffers from temporal incoherence. Extending EG360 to panoramic video depth estimation \cite{songpvdepth, chen2025video} via cross-frame fusions may reduce depth-flickering effects.

% \clearpage
% \input{sec/6_AIstatement}

{
    \small
    \bibliographystyle{ieeenat_fullname}
    \bibliography{main}
}
% \clearpage

\end{document}